## Data Descriptor

# An interactive enhanced driving dataset for autonomous driving

## Authors

Haojie Feng[1†], Xinrui Zhang[1†], Mengjie Tian[1], Peizhi Zhang[1*], Zhuoren Li[1], Junpeng Huang[1], Xiurong Wang[1], Junfan Zhu[3], Jianzhou Wang[4], Dongxiao Yin[2], Lu Xiong[1*]

**Affiliations**

1. School of Automotive Studies, Tongji University, Shanghai, 201804, China
2. Tongji Automotive Design & Research Institute Co., Ltd., Shanghai, 201804, China
3. University of Chicago, Chicago, IL 60637, USA
4. Faculty of Computer Science, University of New Brunswick, Fredericton, NB E3B 5A3, Canada

[†]These authors contributed equally to this work.

[*]corresponding author(s): Peizhi Zhang (zhangpeizhitom@126.com);
Lu Xiong (xiong_lu@tongji.edu.cn)

## Abstract

Driving interaction data are important for training and evaluating autonomous driving Vision-Language-Action (VLA) models, but existing datasets contain limited dense interaction samples and weak alignment between trajectories, visual inputs, and language annotations. This work presents the Interactive Enhanced Driving Dataset (IEDD), a large-scale interaction-oriented dataset constructed from five naturalistic trajectory datasets: Lyft Level 5, Waymo, nuPlan, INTERACTION, and SIND. IEDD contains 7.31 million ego-centric interaction segments, including 6.66 million multi agents cases, covering head-on, car-following, merging, and crossing interactions. Each segment is associated with trajectory-derived interaction metrics describing interaction intensity and efficiency. Based on these annotations, IEDD-VQA further provides trajectory-reconstructed BEV videos, structured interaction semantics, and multi-turn question-answer pairs. The dataset can support interaction mining, long-tail scenario analysis, VLA instruction tuning, and hierarchical evaluation of perception, behavior description, physical quantification, and counterfactual reasoning.

## Background & Summary

Interactive driving behaviors, such as merging, crossing, car-following, and yielding, are common in natural traffic and are closely related to autonomous driving perception, prediction, and decision-making. Existing studies have shown that automated driving systems still face challenges in complex negotiation scenarios, especially under turning, crossing, or visually demanding conditions[1-3]. With the development of Vision-Language-Action (VLA) models, autonomous driving data are increasingly expected to provide not only trajectories or sensor observations, but also structured semantics and language annotations that can support interaction understanding and reasoning[4-12].

However, interaction-related samples remain sparse and weakly annotated in many public driving datasets. Most naturalistic datasets, such as NGSIM[13], nuScenes[14], Lyft Level 5[15], and the Waymo Open Motion Dataset[16], mainly contain routine driving behaviors, while safety-critical interactions, including congested merging and yielding to emergency vehicles, belong to a rare long-tail distribution[17]. Although large-scale perception datasets such as KITTI[18], BDD100K[19], and ONCE[20] provide valuable visual information, and trajectory datasets provide motion states such as position, velocity, and acceleration, most existing resources still lack explicit language annotations describing driver intentions, interaction context, and causal relationships[21-24]. This weak alignment among trajectories, visual observations, and language semantics limits the ability of VLMs and VLA models to learn interaction-aware driving representations. Building a new large-scale multimodal driving dataset from scratch would require specialized sensor platforms, large-scale data collection, cross-modal synchronization, and labor-intensive annotation. Therefore, enhancing existing public trajectory datasets through scalable, physically grounded interaction mining and multimodal synthesis offers a practical alternative. In this study, we first extract ego-centric interaction scenarios from heterogeneous public trajectory datasets, then quantify the physical interaction process in terms of interaction intensity and traffic efficiency, and finally synthesize complementary BEV visual data and structured language annotations. This process enables the construction of interaction-oriented benchmarks and fine-tuning resources for autonomous driving VLMs and VLA models.

To construct the Interactive Enhanced Driving Dataset (IEDD), which features high interaction density and multimodal alignment, we develop a scalable data production pipeline, as illustrated in Fig. 1. The pipeline contains three closely coupled modules. First, the interaction mining module, shown in Fig. 1A, extracts four representative categories of ego-centric interaction segments from heterogeneous naturalistic driving datasets, including head-on, car-following, merging, and crossing scenarios. Second, the interaction quantification module, shown in Fig. 1B, computes physics-aware indicators to describe the intensity, risk, and efficiency of each interaction process, thereby assigning interpretable physical attribute labels to each segment. Third, the multimodal synthesis module, shown in Fig. 1C, reconstructs BEV videos from real trajectories and generates structured semantic descriptions and multi-turn question-answer pairs. In this way, trajectory, visual, and language modalities are temporally and semantically aligned, forming VLA-compatible training and evaluation data.

The resulting data are intended for reuse in several interaction-oriented autonomous driving tasks. The trajectory-level IEDD annotations can be used to retrieve ego-centric interaction segments, analyze long-tail interaction types, compare interaction intensity and efficiency across scenarios, and construct training or evaluation subsets for prediction, planning, and risk assessment models. The IEDD-VQA subset provides temporally aligned BEV videos, structured semantic records, and multi-turn question-answer pairs, which can be used for VLM/VLA instruction tuning and for evaluating perception, behavior description, physical quantification, and counterfactual reasoning from an interaction perspective. The raw trajectories are derived from previously released public datasets, including Lyft Level 5[15], Waymo Open Motion[16], nuPlan[25], INTERACTION[26], and SIND[27], whereas the interaction mining results, physical metric annotations, BEV reconstructions, structured semantic labels, and VQA samples are released as derived data in IEDD and IEDD-VQA. The accompanying code provides scripts for trajectory preprocessing, interaction segment extraction, BEV rendering, VQA construction, and benchmark evaluation, allowing users to reproduce the data generation process or adapt it to other trajectory datasets. The main contributions of this study are summarized as follows:

- IEDD is constructed from Waymo, nuPlan, Lyft Level 5, INTERACTION, and SIND, containing 7.31 million ego-centric interaction segments and 6.66 million multi agents cases.
- A unified trajectory standardization and interaction mining pipeline is developed to extract head-on, car-following, merging, and crossing scenarios with physics-aware intensity and efficiency annotations.
- IEDD-VQA aligns trajectory-reconstructed BEV videos, structured interaction semantics, and multi-turn QA pairs to support VLA instruction tuning and hierarchical evaluation.

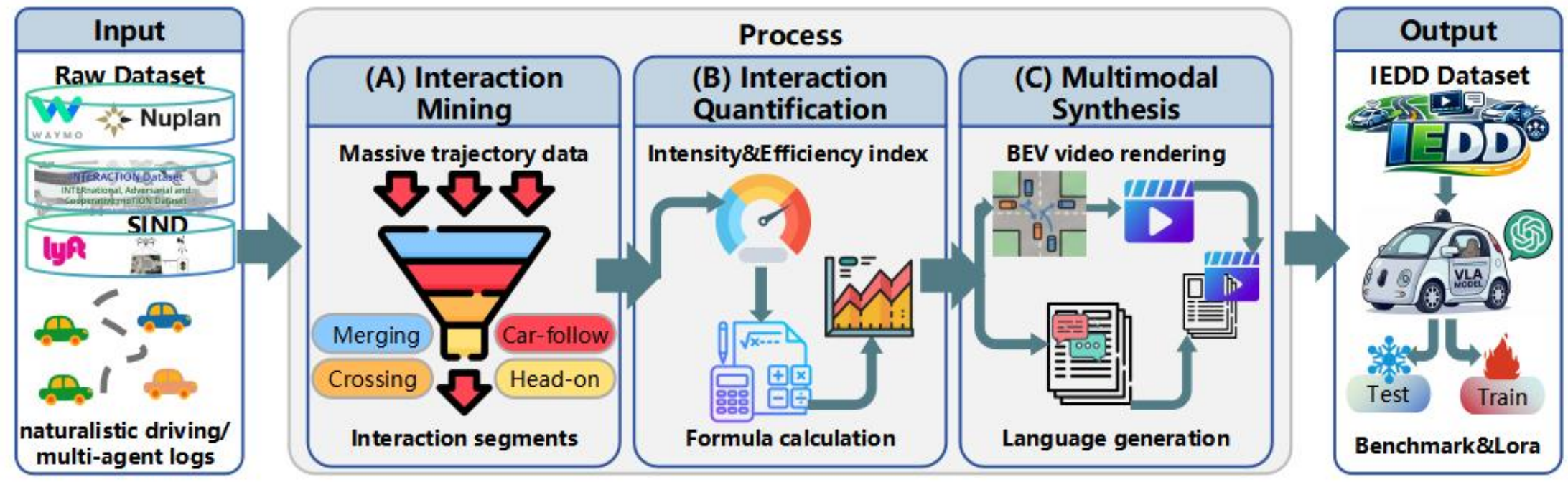

**Fig. 1** Overview of the entire pipeline for generating the IEDD data.

## Related Works

Interaction understanding is essential for autonomous driving systems in complex traffic environments. Autonomous vehicles must not only perceive surrounding objects, but also reason about the intentions, motion dependencies, and negotiation processes among heterogeneous traffic participants, including vehicles, pedestrians, and cyclists. Such interaction-rich scenarios are critical for evaluating decision-making robustness, safety margins, and long-tail risk handling. Recent studies have therefore investigated how to mine interaction scenarios from naturalistic driving data, how to construct interaction-oriented datasets, and how to enrich driving data with semantic and language annotations for vision-language-action learning. This section reviews the related work from three perspectives: interaction scenario mining and quantification, autonomous driving interaction datasets, and VLA-oriented semantic data construction.

*Interaction Scenario Mining and Quantification Methods.* Autonomous vehicles are frequently required to engage in unstructured negotiations with diverse traffic participants in complex urban environments. These interaction events constitute valuable long-tail scenarios for testing safety boundaries and improving the robustness of decision-making models. Due to the high marginal cost of collecting dense interaction scenarios in the real world, early studies mainly explored scenario reconstruction and synthetic generation. Although simulation-based synthesis has been used to augment data volume[28], synthetic datasets often fail to capture the randomness, negotiation characteristics, and physical dynamics inherent in real-world human driving[29]. This leads to a non-negligible Sim2Real distribution gap, and in some cases, the introduction of synthetic data may even degrade model performance in real traffic environments.

In contrast, extracting real interaction segments directly from naturalistic driving data has become a practical solution that balances authenticity and acquisition cost. In recent years, researchers have proposed various extraction strategies based on physical metrics and optimization algorithms. For instance, MSAA formulates driving interaction as a global optimization problem[30], using the minimum change in acceleration as a core metric to identify valid interaction segments. PODAR introduces potential field theory[31] to measure interaction risk by predicting the future trajectories of surrounding vehicles and estimating potential collision kinetic energy. VistaScenario further enables refined scenario measurement and comparison through graph computation trees and Dynamic Time Warping (DTW)[32], while other studies employ interaction primitives to automatically extract multi-vehicle interaction patterns through unsupervised clustering. In addition, several studies model the complexity of dynamic scenarios as the aggregation of pairwise interactions, where interaction intensity is calculated based on kinematic quantities such as encounter angle, relative distance, and relative velocity[33-36]. Although these methods have shown effectiveness in specific scenarios, the field still lacks a general and standardized metric that can consistently quantify interaction intensity while supporting higher-level semantic interpretation[37,38].

*Current Status of Autonomous Driving Interaction Datasets.* In recent years, large-scale autonomous driving datasets, represented by Lyft Level 5[15], Waymo Open Motion Dataset[16] and nuPlan[25], have significantly advanced environment perception and trajectory prediction research. However, these datasets were primarily designed for object-level perception tasks, such as detection, tracking, and semantic segmentation. Despite their large scale, they still exhibit clear limitations in the coverage density and semantic representation of interactive driving behaviors[39-42]. Since routine behaviors, such as driving straight or following at a safe distance, dominate naturalistic driving data, critical interaction scenarios, including forced merging at congested ramps, yielding to emergency vehicles, and negotiation at unprotected intersections, remain sparse and belong to a long-tail distribution[17]. This sparsity makes it difficult for data-driven models to sufficiently learn strategies for handling safety-critical interactions. Moreover, most existing data logs lack explicit annotations of drivers' decision-making rationales and interaction intentions, which limits their interpretability and reasoning value.

Datasets specifically designed for interaction research remain limited. The INTERACTION dataset is one of the few pioneering datasets focused on negotiation scenarios[26], but its overall scale and geographic diversity are relatively restricted. InterHub further attempts to extract interaction segments from naturalistic driving data to enrich training samples[30], yet it lacks ego-centric interaction measurement and systematic physical evaluation. To overcome the limitations of relying on a single data source, this study adopts a cross-domain data fusion and refinement strategy. We establish standardized data adaptation interfaces to integrate five heterogeneous datasets, including Lyft Level 5[15], Waymo Open Motion[16], nuPlan[25], INTERACTION[26], and SIND[27]. By mapping raw trajectories with different sensor configurations, geographical contexts, and traffic rules into a unified spatio-temporal representation space, the proposed framework inherits the scale advantages of existing datasets while alleviating the sparsity of interaction samples through targeted mining algorithms. This strategy enables an order-of-magnitude expansion in the number of interaction scenarios and provides a more diverse foundation for interaction-oriented autonomous driving research.

*VLA Datasets and Semantic Construction for Autonomous Driving.* With the development of Embodied AI, autonomous driving research is shifting from traditional modular architectures toward the end-to-end Vision-Language-Action (VLA) paradigm. In this paradigm, high-quality language labels are no longer merely descriptive annotations; instead, they serve as semantic bridges that connect perceptual inputs with decision-making outputs. Such annotations provide crucial supervisory signals for the closed-loop learning process of "interaction understanding—logical reasoning—action execution". However, constructing large-scale and reliable VLA datasets still faces a fundamental trade-off between the high cost of manual annotation and the hallucination noise introduced by automated generation. Existing language-data construction strategies can be broadly summarized into three paths. First, expert knowledge-based manual annotation emphasizes the authenticity, interpretability, and naturalness of language annotations. For example, BDD-X asks annotators familiar with driving rules to act as "driving instructors" and explain driving behaviors in videos[43], while DRAMA requires all question-answer pairs and captions to be manually written by human annotators[44]. Although manual annotation has a natural advantage in logical consistency, its scalability is severely constrained by time and labor costs. Second, structured data-based template filling improves scalability by leveraging object attributes, lane topology, and scene graphs. NuScenes-QA models driving scenes as Scene Graphs[45] and automatically generates large-scale QA pairs through predefined templates, while Reason2Drive adopts a "structured data template filling + GPT-4 rewriting" scheme to improve linguistic diversity while maintaining factual consistency[46]. This strategy achieves a compromise between scale and reliability, but its dependence on predefined templates makes it difficult to capture complex and dynamic interaction logic. Third, hybrid pipelines that combine large-model generation with strict verification have become a mainstream trend in VLA dataset construction. These methods aim to exploit the reasoning capabilities of LLMs while maintaining structural rigor through rule constraints or human verification. For example, building upon Chain-of-Thought (CoT) reasoning[47], works like DriveMLLM[48] and Impromptu VLA[49] employ a two-stage process where models generate intermediate visual reasoning before summarizing it into QA pairs, effectively extending language supervision from simple perception to causal reasoning. DriveLM introduces graph-structured QA[21], connecting the "perception-prediction-planning" process through logical dependencies, while DriveAction adopts an LLM-based structured scene analysis pipeline with dual human verification[50] and temporal window screening to ensure alignment with human driving intentions.

Inspired by these works, this paper argues that generic scene descriptions are insufficient for modeling the negotiation demands of autonomous driving in complex interactive scenarios. Unlike existing datasets that focus mainly on static scene understanding or single-agent behavior description, we reconstruct the language supervision system around "interactive reasoning". By drawing on the logical-chain design of DriveLM and the verification mechanism of DriveAction[21,50], we establish a structured mapping among physical-world interaction causality, ego-vehicle intentions, and the linguistic logic of VLA models. This design directly addresses the limitations of existing VLA datasets in representing multi-vehicle negotiation and long-horizon interaction reasoning, leading to an interaction-enhanced dataset with high logical consistency[51-56].

# Methods

This section describes the data production pipeline used to construct IEDD and IEDD-VQA. The pipeline follows the roadmap shown in Fig. 1 and includes trajectory preprocessing and scenario slicing, interaction intensity and efficiency quantification, multimodal BEV-language data synthesis, and dataset statistics.

## Naturalistic driving trajectory preprocessing and scenario slicing

We use a standardized pipeline to mine and classify interaction scenarios from naturalistic trajectory data. Taking time-series vehicle motion states as input, this method employs four cascaded steps—trajectory preprocessing, spatio-temporal intersection detection, two-vehicle interaction classification, and multi agents aggregation—to output a set of interaction events containing precise time windows and semantic types.

First, the raw trajectory data are cleaned and standardized. All vehicle trajectories are resampled to a time resolution of 0.1 s, and invalid zero-point data are removed. To address the issue of sudden heading angle fluctuations caused by low-speed driving or localization noise, this study calculates initial headings using the finite difference method. Subsequently, the angular sequence is decomposed into continuous signals in the angular domain and smoothed using a symmetric sliding window. Finally, the signals are mapped back to the standard angular interval, ensuring the continuity and reliability of vehicle kinematic information.

Building on this, the algorithm searches for potential trajectory intersections in the spatio-temporal dimension to identify interaction candidates. By defining spatial distance $D_{search}$ and time difference thresholds $T_{search}$, an intersection is deemed valid—and its geometric center and temporal midpoint are recorded—whenever the trajectory points of two vehicles simultaneously satisfy these constraints within a spatio-temporal neighborhood. To improve efficiency, we use a time-axis double-pointer sliding window that restricts pairwise searches to a local temporal neighborhood.

Subsequently, based on the statistical characteristics of the intersection point set, this paper employs a two-stage classification logic to subdivide interaction types. The first stage targets Car-following behavior: when the number of trajectory overlap points is significant and the median heading difference is below a specific threshold, the interaction is classified as car-following, and the time interval fully covering the intersection point set is defined as the event window. If the car-following criteria are not met, the algorithm proceeds to the second stage, where events are categorized into Merging, Crossing, and Head-on based on the relative heading angle of the two vehicles at the moment closest to the intersection. Fig. 2 illustrates the spatial topological features of these different interaction types. For Merging and Crossing scenarios, the algorithm applies heading angle difference thresholds $\theta_{merge}$ and $\theta_{cross}$ to identify and extract conflict areas (indicated by yellow circles in the figure). Meanwhile, multi agents interactions are defined as complex topologies centered on the ego-vehicle, encompassing all surrounding traffic participants with spatio-temporal overlaps. Finally, interaction time windows $T_{window}$ are uniformly extracted to fully capture the driver's negotiation and decision-making processes.

Finally, to capture the chain reactions within complex traffic flows, the algorithm performs multi agents interaction aggregation. Vehicles involved in multiple interaction types simultaneously within the same scene (e.g., concurrent car-following and lane-changing) are identified as anchor points. All associated traffic participants are then recursively merged to form a unified Multi agents Group. The effective time window for this group is defined as the global union of the time intervals of all constituent interaction events. This approach ensures the spatio-temporal and semantic integrity of complex scenarios while preventing the fragmentation of interactive behaviors.

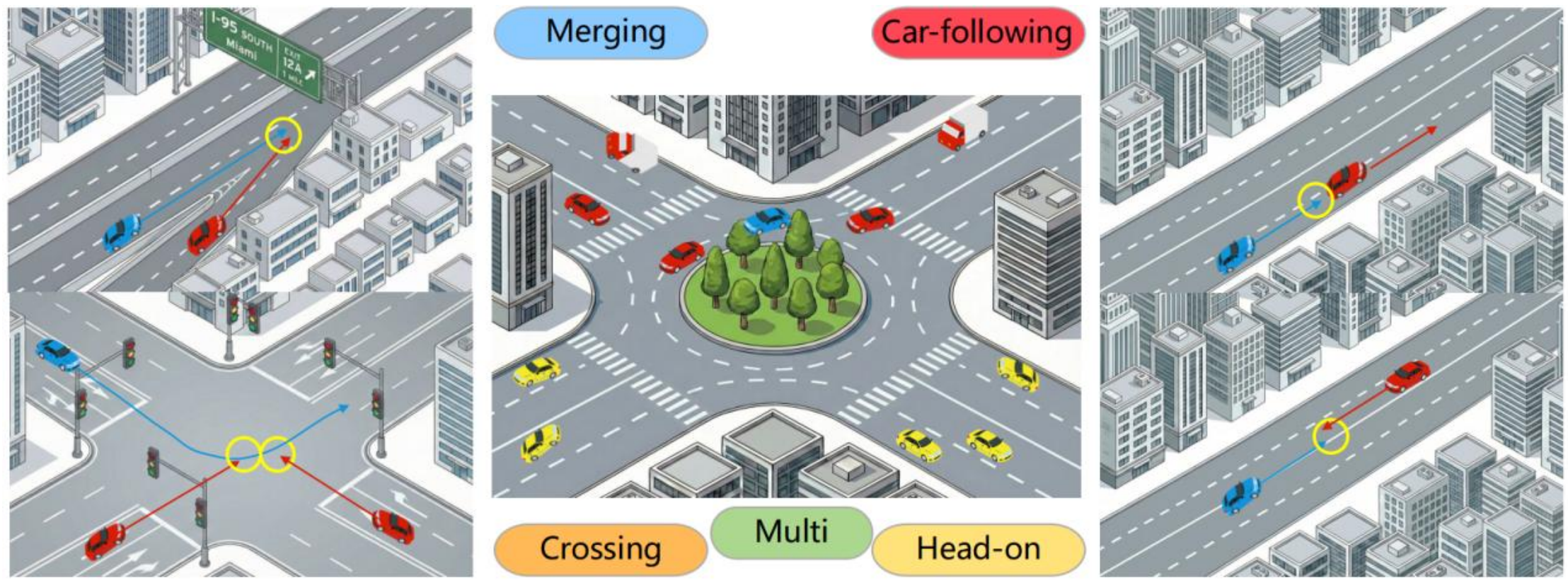

**Fig. 2** Schematic illustration of the geometric classification for typical interaction scenarios.

## Interaction metric system based on intensity and efficiency

To evaluate the fine-grained intensity and efficiency of interaction processes, this study establishes a quantitative evaluation system based on the stochastic processes of interaction. We model vehicle-to-vehicle interactions as continuous time-series processes. By defining interaction intensity and interaction efficiency, we provide a multi-level characterization of interactive behaviors through the dual dimensions of "process risk evolution" and "decision execution quality."

(1) ***Interactive state space modeling***

To transcend treating interaction segments as mere collections of isolated trajectory points, we formally model them as observed realizations of a multi agents stochastic state process. Within this framework, each extracted trajectory segment operates as a single sample path of the continuous process X(t). Crucially, X(t) is not deployed as a parametric predictive model; thus, no rigid Markovian, Gaussian, or specific transition-distribution assumptions are imposed. Rather, it serves as a foundational and unified state representation, from which all subsequent interaction events, quantitative physical indicators, and semantic language labels are systematically derived.

For $n$ traffic participants in the scene, the state vector $S_i(t)$ of the i-th participant at time t is defined as:

$$X(t)=\{S_1(t),S_2(t),\ldots,S_n(t)\} \tag{1}$$

$$S_i(t)=[x_i(t),y_i(t),\varphi_i(t),v_i(t),a_i(t)]^T \tag{2}$$

where $(x_i,\ y_i)$ represents the planar coordinates, while $\varphi_i, v_i$, and $a_i$ denote the heading angle, velocity, and acceleration, respectively. To capture valid interaction events, we define triggering thresholds based on spatio-temporal neighborhoods: an interactive state is identified when two vehicles simultaneously satisfy both the spatial distance threshold $d_{inter}$ and the temporal difference threshold $t_{inter}$ for interaction triggering.

Specifically, the discrete state sequence extracted from $X(t)$ serves as the fundamental quantitative basis for both physical measurement and semantic generation in our pipeline. On the one hand, the state variables encoded in $X(t)$ are fed directly into the calculation of the dynamic interaction intensity. The kinematic parameters determine the pose adjustment metric, while the planar coordinates and heading angles dynamically construct the spatio-temporal relative distances, risk gradients, and interaction potential fields at each respective time step. On the other hand, the continuous time-series evolution within $X(t)$ provides the essential physical evidence for extracting behavioral atoms. By continuously evaluating the instantaneous fluctuations of variables within $S_i(t)$ against predefined kinematic thresholds, the numerical trajectory states are strictly discretized into symbolic semantic actions. The temporal continuity preserved in the resampled sequence further allows these instantaneous atoms to be aggregated into stable behavioral chains, effectively bridging the gap between low-level kinematic data and high-level linguistic reasoning.

(2) ***Dynamic interaction intensity evaluation metrics***

The interaction intensity metric $Q_i(t)$ aims to quantify the instantaneous conflict pressure and the intensity of response maneuvers faced by vehicles during the interaction process. This metric is formulated as a weighted coupling of three physical components: pose adjustment, risk gradient, and environmental potential energy:

$$Q_i(t)=w_sQ_{si}(t)+w_rQ_{ri}(t)+w_pQ_{pi}(t) \tag{3}$$

where $w_s$ , $w_r$ , $w_p$ denote the normalized weight coefficients. Considering the significant variations in drivers' risk perception emphasis across different types of interactive scenarios, this paper adopts a category-adaptive weight allocation strategy. Specifically:

Merging Scenarios: Since merging maneuvers are highly dependent on the assessment of lane space and the potential fields of surrounding vehicles, a higher weight is assigned to the potential energy term.

Crossing Scenarios: In crossing interactions, the temporal evolution of collision risk is relatively intense; thus, the weight of the risk variation term is significantly increased to capture critical conflict moments.

Head-on Scenarios: These scenarios are typically accompanied by extremely high relative speeds. As drastic fluctuations in Time-to-Collision (TTC) and Post-Encroachment Time (PET) serve as the core indicators of danger, the risk term holds a dominant position.

This differentiated weighting scheme ensures that the quantification metrics can sensitively capture the core risk characteristics within each specific interactive pattern.

The physical definitions of each component are as follows:

(2.1) ***Pose Adjustment Metric*** $\boldsymbol{Q_{si}(t)}$ ***:*** This metric quantifies the magnitude of the kinematic response executed by the vehicle to mitigate potential conflicts. It is calculated using the normalized rate of change of velocity $\Delta v_i$ and acceleration $a_i$:

$$Q_{si}(t)=\frac{|\Delta v_i(t)|}{v_{lim}}+\beta_Q\frac{|a_i(t)|}{a_{lim}} \tag{4}$$

where $\beta_Q$ is the acceleration adjustment weight, $v_{lim}$ denotes the vehicle speed limit, and $a_{lim}$ represents the maximum acceleration capability. A higher value of this metric indicates that the vehicle has executed more aggressive acceleration/deceleration or maneuvering adjustments.

(2.2) ***Risk Variation Metric*** $\boldsymbol{Q_{ri}(t)}$ ***:*** This reflects the temporal evolution trend of collision risk. In contrast to static indicators that rely solely on a single instantaneous state, we introduce the time derivatives of TTC and PET to characterize the imminence of risk approach:

$$Q_{ri}(t)=\left(\frac{1}{TTC(t)}-\frac{1}{TTC(t-\Delta t)}\right)+\gamma_Q\left(\frac{1}{PET(t)}-\frac{1}{PET(t-\Delta t)}\right) \tag{5}$$

where $\gamma_Q$ is the sensitivity factor for PET variation, and the difference between the two terms represents the amount of change over adjacent time steps. This formulation effectively captures critical moments of "sharp risk escalation."

(2.3) ***Interactive Potential Field Metric*** $\boldsymbol{Q_{pi}(t)}$***:*** This metric constructs a dynamic risk distribution based on the Artificial Potential Field (APF) method. For all surrounding neighboring vehicles, the superimposed potential energy exerted on the ego vehicle is calculated as:

$$Q_{pi}(t)=\sum_{j\epsilon N_i(t)}\omega_{dir}(\varphi_{ij})exp[-(\frac{d_{ij}}{d_0})^2][1+\kappa_v\sigma(\frac{v_{ij}^{||}}{v_0})] \tag{6}$$

where $N_i(t)$ is the set of traffic participants surrounding the ego vehicle at time t, $\omega_{dir}(\cdot)$ is the direction weighting function, $\varphi_{ij}$ represents the relative azimuth, $d_{ij}$ is the relative distance between the two vehicles, $d_0$ is the potential field distance factor, $v_0$ is the relative velocity normalization factor, $\kappa_v$ is the velocity sensitivity coefficient, $\sigma(\cdot)$ is the sigmoid compression function, and $v_{ij}^{||}$ is the relative approach velocity between the two vehicles. Through the direction weighting function, the model assigns higher potential energy weights to the region in front of the ego vehicle, aligning with the human driver's perceptual characteristic of being more sensitive to frontal risks.

The interaction intensity metrics primarily focus on quantifying the dynamic risk and interaction pressure during the driving process. However, in real-world traffic scenarios, optimal driving strategies involve more than mere risk avoidance; they must also achieve efficient and smooth traversal under the premise of ensuring safety. Relying solely on intensity metrics makes it difficult to comprehensively evaluate the overall quality of an interactive behavior. Therefore, to establish a complete closed-loop evaluation of interaction behavior, it is necessary to introduce a result-oriented efficiency measurement dimension.

It should be noted that the APF term in Eq. (6) is used in IEDD as a scalable post-hoc descriptor of interaction pressure rather than as a closed-loop motion planner. Classical APF methods are effective in representing distance-, direction-, and velocity-dependent risk distributions, but they may be insufficient for fully capturing strategic negotiation in highly interactive scenarios, such as forced merging, roundabout negotiation, and multi agents yielding. In these cases, the risk perceived by the ego vehicle depends not only on geometric proximity, but also on the expected responses and priorities of surrounding agents.

A possible extension is to formulate a game-aware dynamic potential field, where the potential weight is modulated by an inferred negotiation coefficient derived from leader-follower roles, right-of-way relations, or predicted yielding/aggressive responses. Recent studies have shown that integrating Stackelberg games with dynamic potential fields can improve the modeling of interactive decision-making under uncertainty[57]. Inspired by this direction, future versions of IEDD may replace the fixed APF weighting with response-aware potential coefficients estimated from inverse game-theoretic reasoning or world-model rollouts. This would allow the interaction intensity metric to better reflect not only instantaneous physical risk, but also the strategic negotiation process among traffic participants.

(3) ***Comprehensive interaction efficiency metrics***

In addition to the risk intensity during the process, evaluating the performance of interaction strategies also necessitates the consideration of the final traversal quality. This paper develops a comprehensive efficiency metric, formulated as the product of three independent dimensions—path, time, and smoothness—with a normalized range of [0,1]:

$$E_i = E_{pi} \cdot E_{ti} \cdot E_{si} \tag{7}$$

(3.1) ***Path Consistency Metric*** $\boldsymbol{E_{pi}}$***:*** This metric quantifies the geometric alignment between the actual trajectory and the reference path:

$$E_{pi} = \frac{d_{ref}}{d_{actual}} = \frac{\sqrt{(x_{i0}-x_{iT})^2+(y_{i0}-y_{iT})^2}}{\int_0^T v_i(t)\,dt} \tag{8}$$

where the numerator represents the reference path length, and $d_{ref}$ denotes the Euclidean distance between the starting and ending points. The denominator is the actual trajectory length. This metric reflects the vehicle's capability to adhere to the optimal trajectory during high-curvature maneuvers.

(3.2) ***Time Consistency Metric*** $\boldsymbol{E_{ti}}$***:*** This evaluates the delay penalty based on the vehicle's desired speed:

$$E_{ti} = exp(-\alpha_E \frac{T_{delay}}{T_{free}}) \tag{9}$$

where $T_{delay}$ represents the cumulative time loss caused by interaction, $T_{free}$ denotes the free-flow time, and $\alpha_E$ is the time sensitivity coefficient. In interactive scenarios, strategies capable of concluding the game with minimal speed loss are assigned higher scores.

(3.3) ***Driving Smoothness Metric*** $\boldsymbol{E_{si}}$***:*** Passenger comfort is evaluated by calculating the standard deviation of acceleration within the interaction window:

$$E_{si} = exp(-\beta_E \frac{\sigma_a}{a_{normal}}) \tag{10}$$

where $\sigma_a$ is the standard deviation of acceleration, $a_{normal}$ denotes the acceleration normalization benchmark, and $\beta_E$ represents the smoothness sensitivity coefficient. Drastic fluctuations in acceleration and deceleration result in an exponential decay of the score for this metric.

To ensure the reproducibility of the data mining and quantification processes, all key physical parameters and threshold presets involved in the interactive scenario slicing and the calculation of interaction intensity and efficiency are summarized in Table 1.

Fig. 3 illustrates the physical significance of the aforementioned quantitative metrics. As the spatio-temporal distance between the two vehicles decreases (from t=0.8s to t=2.0s), the color at the center of the environmental potential energy heatmap gradually intensifies, corresponding to a significant rise in the interaction intensity curve (as shown in the bottom-left of Fig. 3). Concurrently, the bar chart in the bottom-right quantifies the traversal efficiency of different vehicles after the game-theoretic interaction, effectively distinguishing the cost differences between aggressive and conservative strategies.

In summary, the interaction intensity metric is employed to precisely locate high-risk "critical scenarios" within the long-tail distribution, while the comprehensive interaction efficiency metric serves as a reward signal to evaluate interactive traversal efficiency based on safety boundaries.

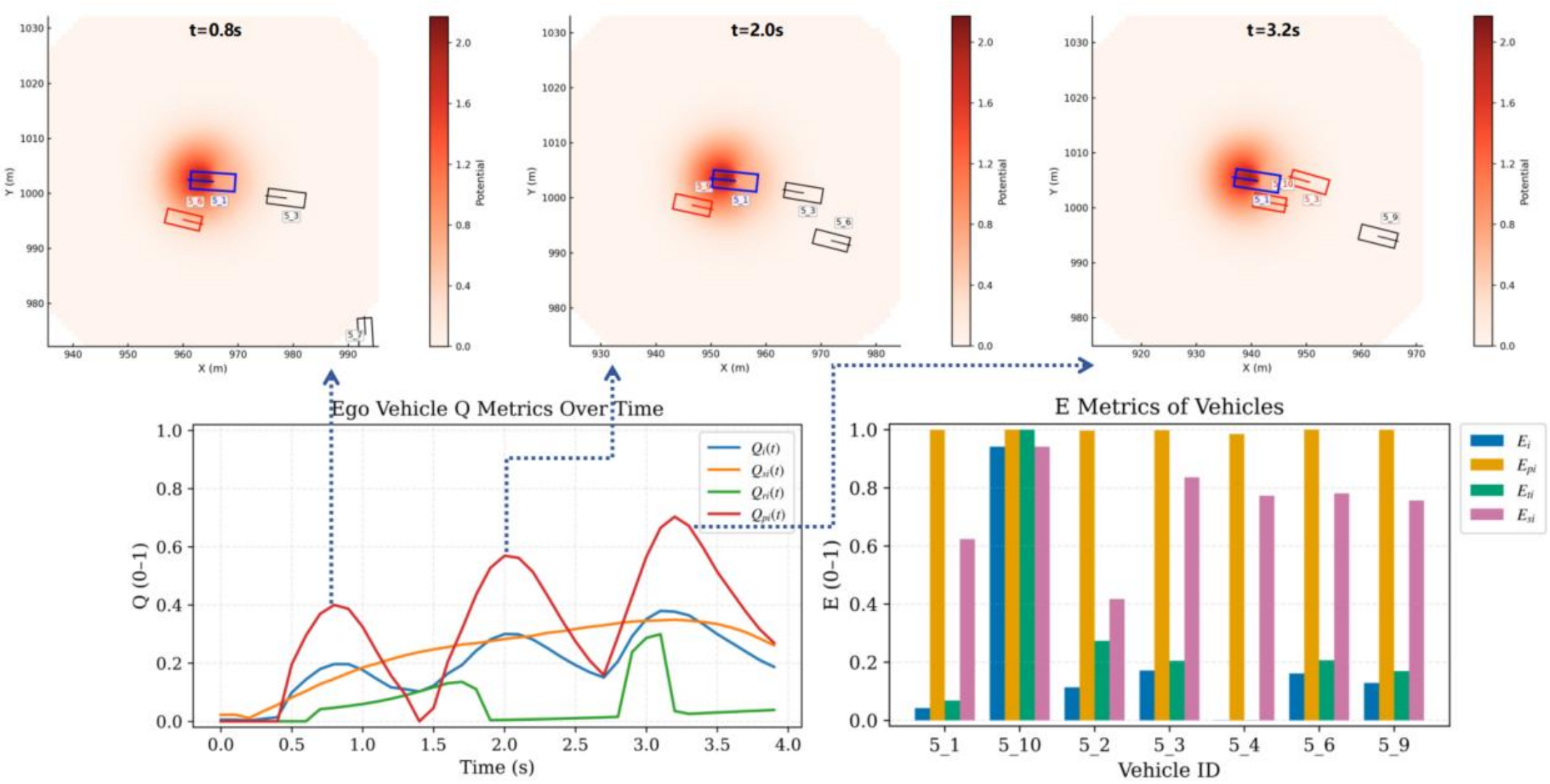

**Fig. 3** Spatio-temporal visualization of dynamic interaction metrics.

| Category | Symbol | Description / Physical Meaning | Value |
|---|---|---|---|
| Interaction Mining | $D_{search}$ | Spatial distance threshold for intersection search | 2m |
| | $T_{search}$ | Temporal difference threshold for intersection search | 3s |
| | $T_{window}$ | Time window for interaction event extraction | 5s |
| | $\theta_{merge}$ | Heading angle threshold for Merging scenarios | 30° |
| | $\theta_{cross}$ | Heading angle threshold for Crossing scenarios | 160° |

| State Definition | $d_{inter}$ | Interaction trigger distance threshold | 50m |
|---|---|---|---|
| | $t_{inter}$ | Interaction trigger time threshold | 3s |
| Intensity Weights | $w_s$, $w_r$, $w_p$ | Weight coefficients for Merging scenarios | 0.25, 0.35, 0.40 |
| | $w_s$, $w_r$, $w_p$ | Weight coefficients for Crossing scenarios | 0.20, 0.55, 0.25 |
| | $w_s$, $w_r$, $w_p$ | Weight coefficients for Head-on scenarios | 0.15, 0.65, 0.20 |
| Intensity Parameters | $\beta_Q$ | Weight for acceleration adjustment | 0.4 |
| | $v_{lim}$ | Speed limit normalization factor | 20m/s |
| | $a_{lim}$ | Maximum acceleration capacity factor | $3m/s^2$ |
| | $\gamma_Q$ | Sensitivity factor for PET variation | 0.4 |
| Potential Field | $d_0$ | Effective distance factor for potential field | 8 |
| | $v_0$ | Relative velocity normalization factor | 5 |
| | $\kappa_v$ | Velocity sensitivity coefficient | 1 |
| Efficiency Metrics | $\alpha_E$ | Sensitivity coefficient for time efficiency | 0.8 |
| | $a_{normal}$ | Acceleration normalization baseline | $2m/s^2$ |
| | $\beta_E$ | Sensitivity coefficient for driving smoothness | 1.2 |

**Table 1.** Summary of Key Parameters for Interaction Mining and Quantification.

## Multimodal interaction instruction data generation pipeline

To address the deficiencies of current autonomous driving datasets in VLA alignment and to support end-to-end learning of complex game-theoretic behaviors, this study establishes a scalable data synthesis pipeline. Taking the interaction snippets extracted in "Naturalistic driving trajectory preprocessing and scenario slicing" and the quantitative metrics calculated in "Interaction metric system based on intensity and efficiency" as core inputs, the framework aims to generate triplet samples where visual features, structured semantics, and natural language descriptions are highly coupled. This section details the mapping mechanism from continuous trajectories to discrete semantics, the rule-constrained language generation strategy, and the implementation of cross-modal spatio-temporal alignment.

(1) ***Structured semantic representation and behavior serialization***

To enable VLMs to comprehend continuous vehicle kinematic characteristics, trajectory data are abstracted into symbolic semantic units, forming a structured semantic set that comprises agent behavior chains, interactive relationships, and metadata.

Behavioral Atom Recognition and Temporal Compression: The continuous state space of vehicles is discretized into sequences of "behavioral atoms." First, instantaneous states are determined for each frame based on physical features such as velocity, acceleration, and yaw rate. Subsequently, to eliminate short-term fluctuations caused by sensor noise and to extract high-level semantics, a temporal aggregation strategy is employed to merge consecutive homogeneous atoms into compact "behavioral chains." This process explicitly records the start and end time windows for each action, providing precise temporal evidence for subsequent linguistic descriptions.

Interaction Relationship Modeling and Phase Partitioning: Since the behavior of a single

agent cannot fully characterize the game-theoretic process, we further construct models of interactive relationships between agents. Using the peak interaction intensity as a time anchor, the interaction process is divided into three evolutionary phases: "Approach," "Interaction," and "Outcome." Within each phase, relative bearing, right-of-way, and specific interaction types are explicitly extracted to endow the data with causal logic.

Encapsulation of Explainable Metadata: To ensure data consistency, the role labels of participating agents and the quantized interaction intensity levels are encapsulated as metadata. These metadata serve as semantic constraints, ensuring that the subsequently generated linguistic descriptions remain strictly faithful to physical ground truth and avoid ambiguous expressions.

Through the aforementioned steps, continuous and complex trajectory data are successfully parsed into discrete behavioral atoms and interaction relationships with explicit semantics. These structured semantic representations serve as a bridge connecting the physical world with the linguistic space. Building upon this structured foundation, this section will elaborate on the construction of a rule-driven generation mechanism to transform these semantic insights into high-quality, hallucination-free natural language supervision signals for guiding the training of VLA models.

(2) ***Rule-driven language supervision generation mechanism***

To eliminate potential "hallucination" phenomena during the large-scale model generation process, this paper proposes a controlled text generation method driven by structured semantics. Language annotations are no longer freely generated text; instead, they are constructed based on predefined semantic slots and logical templates.

(2.1) ***Intensity-Aware Semantic Mapping:*** The calculated continuous interaction intensity metrics are mapped onto discrete linguistic modifiers, and interpretability is enhanced through the introduction of "evidence phrases." Specifically, the system automatically extracts micro-behavioral changes (such as sharp deceleration or heading adjustments) near peak intensity moments and converts them into specific textual descriptions.

(2.2) ***Multi-Level Instruction Construction:*** To accommodate various training objectives, three complementary text formats are generated: global summaries outlining interacting objects and types; standardized action chains representing the interaction process; and multi-turn question-answering (QA) instructions encompassing scene perception, intent reasoning, and policy recommendations. All QA answers are strictly extracted from structured semantics, thereby enhancing the model's generalization capabilities while ensuring factual accuracy.

While high-quality linguistic instructions empower the model to understand interaction logic, another core pillar of VLA models is visual perception. A single linguistic modality is insufficient to support the model's intuitive understanding of complex spatial relationships. To construct multimodal samples with strict image-text correspondence, visual inputs must be generated that are frame-level aligned—both spatially and temporally—with the aforementioned trajectory ground truths. The following section details the trajectory-driven BEV video rendering and its synchronization strategy with linguistic instructions.

(3) ***Trajectory-driven BEV rendering and spatio-temporal alignment***

Visual inputs utilize BEV videos reconstructed from real-world trajectories, aiming to provide the model with clear global spatial awareness and ensure strict alignment between vision and language. The choice to employ BEV rather than front-view or surround-view camera images as the primary visual modality is based on dual considerations of pipeline versatility and the completeness of interaction reasoning. First, to overcome the high heterogeneity of source data sources—specifically, the significant differences in sensor configurations across datasets (such as INTERACTION, SIND, and Waymo) and the fact that many open-source trajectory-based datasets do not include synchronized surround-view imagery—we adopt a BEV rendering scheme based on high-precision trajectory reconstruction. This strategy eliminates dependence on specific hardware, ensuring that the data generation pipeline can seamlessly migrate to any trajectory dataset, thereby maximizing the scale expansion potential of IEDD-VQA.

Second, in complex interaction scenarios, first-person perspectives often face field-of-view limitations and object occlusions, making it difficult to fully capture the global dynamics of multi agents games. In contrast, the BEV perspective provides an unoccluded spatial representation from a "god's-eye view," explicitly presenting the road topology and relative positional relationships of surrounding vehicles. This omnidirectional perception is crucial for VLA models to understand global interaction logic, predict potential conflicts, and plan long-term trajectories, serving as key auxiliary information that a single front-view perspective cannot replace.

(3.1) ***Ego-Centric Standardized Rendering:*** To normalize model inputs, all trajectories are transformed into a unified BEV coordinate system, with the origin set to the ego vehicle's pose at the peak interaction moment. Dynamic vehicles are rendered as oriented geometries, with short-term historical trajectories overlaid to provide motion cues.

(3.2) ***Strict Spatio-Temporal Synchronization:*** The quality of multimodal data depends on the alignment precision across modalities. We establish a unified temporal baseline to ensure that BEV video frames correspond strictly to trajectory sampling times. Crucially, critical moments and behavior intervals referenced in the linguistic descriptions are aligned with video frame indices at a pixel-level. This design enables the model to effectively map visual patterns, physical motion, and linguistic logic, facilitating true multimodal reasoning.

The specific data mapping mechanism is illustrated in Fig. 4. The system first transforms continuous trajectory segments into discrete sequences of atomic actions (e.g., 'accelerate', 'turn left', 'yield'). Subsequently, these semantic labels are filled into predefined dialogue templates to generate question-answering pairs that are strictly aligned with the BEV video frames. The right side of Fig. 4 shows an example of the generated JSON data, covering multiple layers of dialogue including perception, description, and reasoning.

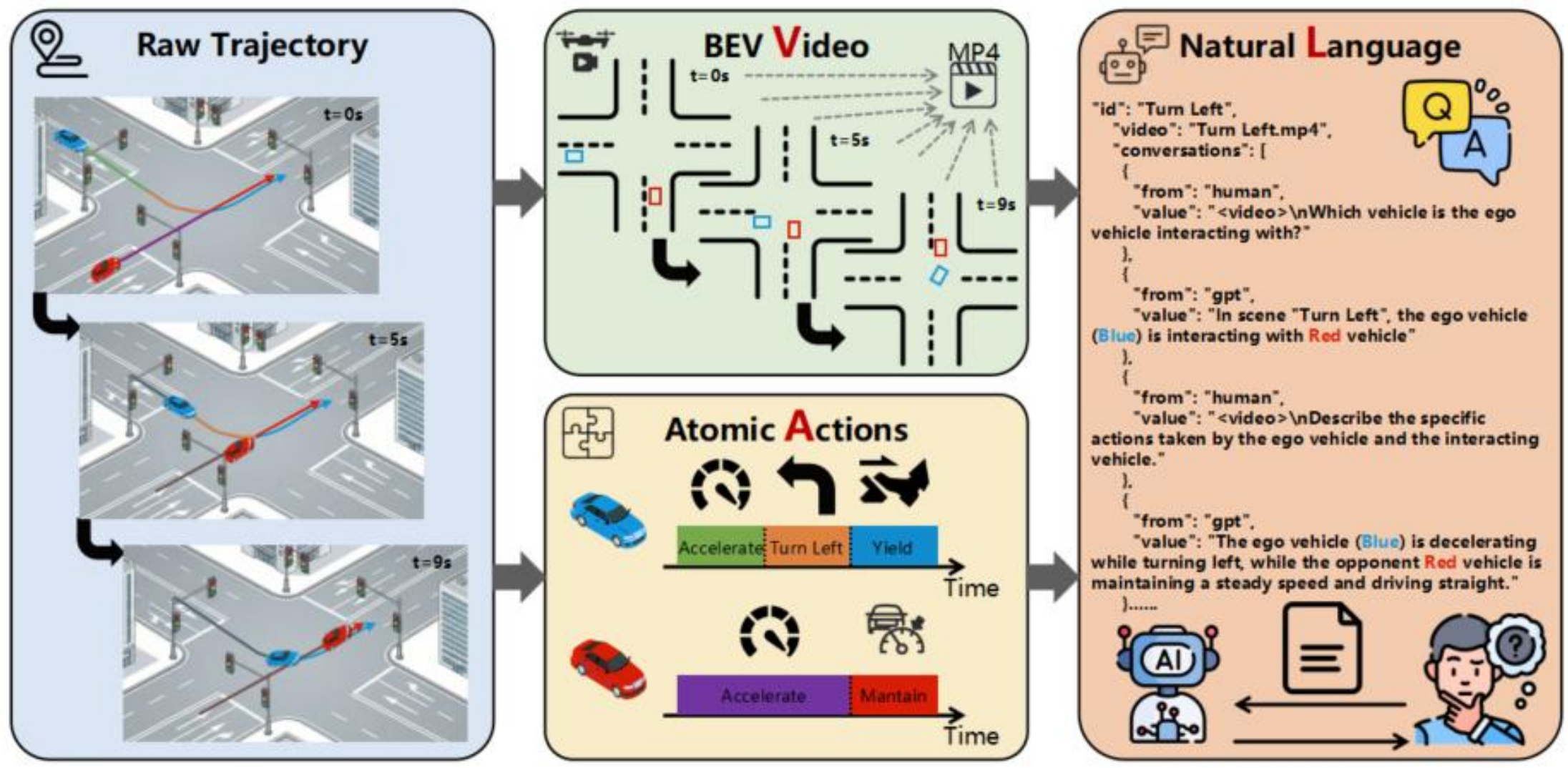

**Fig. 4** Trajectory-driven multimodal data synthesis workflow.

## Source data acquisition

The input data used to construct IEDD were obtained from five previously released naturalistic driving trajectory datasets: Lyft Level 5, Waymo Open Motion Dataset, nuPlan, INTERACTION, and SIND. These datasets were selected because they provide large-scale trajectory records covering diverse traffic environments, road topologies, and interaction patterns. No privately collected raw driving data were used in this study.

- Lyft Level 5[15] was obtained from the publicly released Level 5/Woven Planet prediction dataset resources (https://woven.toyota/en/prediction-dataset). This dataset contains large-scale driving scenes sampled at 10 Hz and was originally designed for trajectory prediction research. In this work, we used the agent trajectory records, scene identifiers, timestamps, and related motion-prediction metadata.
- Waymo Open Motion Dataset (WOMD)[16] was obtained from the Waymo Open Dataset portal (https://waymo.com/open/). WOMD provides motion prediction scenarios collected from multiple cities in the United States. In this work, we used the Scenario Proto records, which contain trajectory sequences sampled at 10 Hz. Other raw sensor modalities, such as camera images and LiDAR point clouds, were not used.
- NuPlan[25] was obtained from the official nuPlan data portal (https://www.nuscenes.org/nuplan). nuPlan is a large-scale closed-loop planning benchmark containing human driving data collected in several cities in the United States and Asia. In this work, only the scenario metadata, trajectory records, and dataset splits were used. Raw sensor data were not included in the construction of IEDD.
- INTERACTION[26] was obtained from the official INTERACTION dataset website (https://interaction-dataset.com/) after completing the dataset access request procedure. INTERACTION provides trajectory records from complex interactive driving scenarios collected in multiple countries and traffic environments. In this work, we used the prediction-oriented trajectory records, including agent positions, headings, velocities, timestamps, and scene identifiers.
- SIND dataset[27] was obtained from the SIND project repository(https://github.com/SOTIF-AVLab/SinD) and through the access procedure specified by the dataset maintainers. The complete dataset can be requested by contacting

the maintainers via the email address provided by the SIND project (13645450063@163.com or 18975505069@163.com). SIND provides drone-based trajectory records at signalized intersections in China. In this work, we used the trajectory-level records, including road-user identifiers, positions, timestamps, velocities, headings, and object categories.

For all five source datasets, only trajectory-level information required for interaction mining was used. Original camera images, LiDAR point clouds, raw videos, and other sensor recordings from the source datasets were not redistributed. To allow users to locate the same input records, each processed IEDD sample retains source dataset provenance information, including the source dataset name, original scene or scenario identifier, agent identifier, frame or timestamp interval, interaction type, and dataset split where available. Users who wish to reproduce the full data generation pipeline should first obtain the corresponding source datasets from their original providers under the respective access procedures and terms of use, and then run the released preprocessing, interaction mining, BEV rendering, and VQA construction scripts.

## Data Records

The IEDD and IEDD-VQA datasets are available at Zenodo[58]. The repository contains trajectory-derived interaction annotations, VQA instruction files, and BEV videos.

The IEDD annotations are organized by source dataset and provided as five compressed packages: IEDD-waymo.zip, IEDD-nuplan.zip, IEDD-lyft.zip, IEDD-interaction.zip, and IEDD-sind.zip. Each package contains CSV files generated from the corresponding source trajectory dataset. The CSV records describe ego-centric interaction segments and include source-dataset provenance, scene or scenario identifiers, interacting agents or multi agents groups, temporal intervals, interaction types, temporal interaction-intensity values, multi agents group information, and vehicle-level efficiency scores.

The IEDD-VQA records are provided as JSON files. IEDD-VQA_train.json contains the training split, covering perception, interaction recognition, behavior description, and physical quantification tasks. IEDD-VQA_test.json contains the benchmark test split, which further includes counterfactual reasoning questions. IEDD-VQA_test_cot.json provides the corresponding test records formatted for Chain-of-Thought prompting. Each JSON entry includes the sample identifier, associated video name and path, source-dataset information, interaction type, participating agents, task level, question-answer pairs, and the corresponding semantic and physical labels.

The BEV videos are provided as MP4 files in IEDD-VQA_train_videos.zip and IEDD-VQA_test_videos.zip. These videos are rendered from standardized trajectory records in a bird's-eye-view perspective and are linked to the JSON records through the video name or path field. Together, the CSV annotations, JSON instruction files, and BEV videos constitute the released IEDD and IEDD-VQA datasets.

## Data Overview

The IEDD dataset is designed to provide multimodal, high-density, and semantically interpretable interactive driving data. Its core content comprises two primary components: a trajectory-based large-scale interaction segment dataset (IEDD) and an instruction tuning dataset for VLA models (IEDD-VQA). This section elaborates on the introduction, statistical characteristics, storage structure, and semantic annotation specifications of the dataset.

(1) ***IEDD scale and statistics***

To validate the advantages of IEDD in interactive scenario coverage, a systematic comparison was conducted between the proposed dataset and the source datasets. The results are summarized in Table 2.

| Dataset | Head-on | Car-following | Merging | Crossing | Two agents | Multi agents | Total |
|---|---|---|---|---|---|---|---|
| INTERACTION | 4642 | 47944 | 15202 | 5267 | 28581 | 44474 | 73055 |
| SIND | 29546 | 36523 | 17589 | 37832 | 263 | 121227 | 121490 |
| nuPlan | 24880 | 44206 | 34464 | 83408 | 21182 | 165776 | 186958 |
| Waymo | 26428 | 781254 | 37675 | 59669 | 442314 | 462712 | 905026 |
| Lyft | 533992 | 714216 | 1284919 | 3494329 | 166296 | 5861160 | 6027456 |
| IEDD | 619488 | 1624143 | 1389849 | 3680505 | 658636 | 6655349 | 7313985 |

**Table 2.** Statistical comparison of interaction types and scales between IEDD and the source datasets.

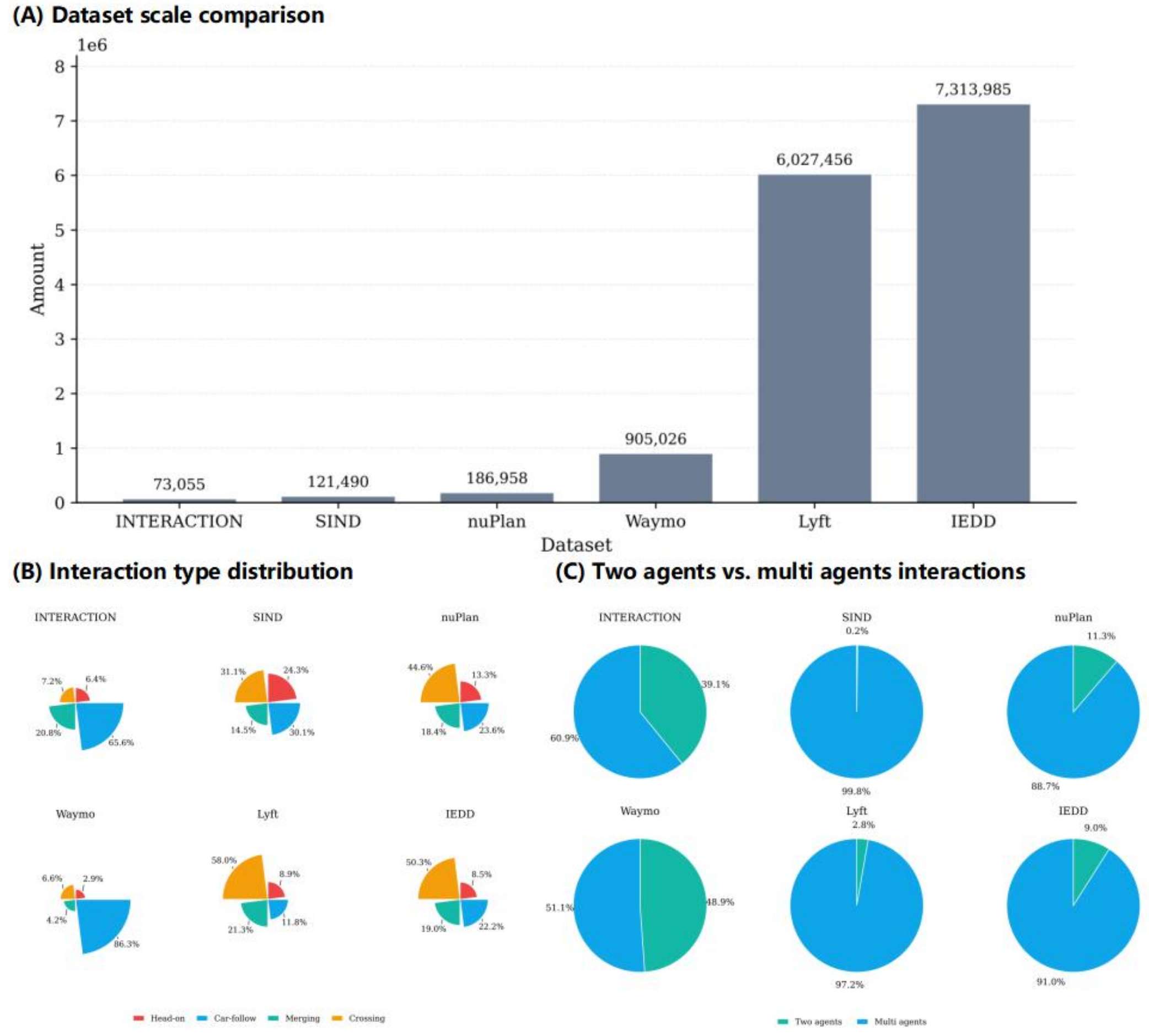

**Fig. 5** Statistical characteristics of IEDD and source datasets.

As illustrated in Table 2 and Fig. 5A, the total volume of the IEDD dataset reaches 7.31 million interaction segments. This scale is comparable to the largest individual source dataset, Lyft, which contains 6.03 million mined interaction segments, and is substantially larger than the other individual datasets such as Waymo, nuPlan, SIND, and INTERACTION. More importantly, the value of IEDD does not lie solely in its absolute scale. By integrating heterogeneous sources into a unified spatio-temporal representation and applying interaction-oriented mining, IEDD provides broader interaction-type coverage, richer multi agents interaction structures, and physically grounded metadata for downstream VLA tasks. Fig. 5B further shows that IEDD is less dominated by simple car-following cases and includes substantial long-tail interaction types, including crossing, merging, and head-on scenarios. Fig. 5C highlights another key property: 91.0% of IEDD samples involve multi agents interactions, in contrast to SIND, where 99.8% of samples are two agents cases.

(2) ***IEDD interaction metadata structure***

The fundamental data units of IEDD are stored in the form of serialized interaction segments, designed to provide a comprehensive description ranging from macro-level scenarios to micro-level dynamics. Table 3 presents the field definitions for the interaction data. Each sample includes a unique Scene ID and a Vehicle Pair index, with the start and end Time Intervals precisely recorded. Beyond basic spatio-temporal attributes, a key feature of IEDD is the integration of quantitative indicators based on physical priors:

Q_i Over Time: The dataset records high-frequency temporal variations of the interaction intensity indicator Q_i, which directly reflect the risk fluctuations and maneuver intensity of the ego vehicle during interaction.

Multi-Vehicle Group: This defines the multi-vehicle interaction topology in complex scenarios, enabling the modeling of cascading interaction behaviors.

E_Veh: The passage efficiency scores of the interacting entities are recorded, providing objective ground truths for assessing decision-making quality.

| Field | Description |
|---|---|
| Scene | Scenario information |
| Vehicle Pair | Interacting vehicle pair |
| Time Interval | Interaction time interval |
| Interaction Type | Interaction type |
| Q_i_Over_Time | Temporal variation of Q_i for the ego vehicle |
| Multi-Vehicle Group | Multi-vehicle interaction group |
| E_Veh | E values of involved vehicles |

**Table 3.** Field definitions of the IEDD dataset.

(3) ***IEDD-VQA data construction***

To support end-to-end learning for Vision-Language-Action (VLA) models, high-value interaction segments were further processed into the IEDD-VQA dataset. This subset is encapsulated in a standardized JSON format, comprising BEV video paths, system prompts, and multi-turn conversations. The dataset is constructed following a "perception-description-quantization-reasoning" CoT logic. The training set specifically includes five major tasks designed around the ground-truth trajectories of the source datasets:

- Perception and Recognition: Identifying the interaction object IDs and their corresponding interaction types relative to the ego vehicle.

• Behavior Description: Requiring the model to describe the specific actions of both vehicles in natural language (e.g., "decelerating to yield" or "maintaining constant speed").

• Physical Quantization: Introducing numerical reasoning tasks regarding interaction intensity values and efficiency scores through a QA format.

•System Prompts: Explicitly defining that the VLM must reason from a BEV perspective based on four predefined interaction primitives.

Notably, the test set introduces a "counterfactual reasoning" task that extends beyond the training set. Fig. 6 illustrates the four-level task structure of IEDD-VQA. In addition to fundamental perception and description, the yellow region in the bottom right of Fig. 6 (L4 Reasoning) showcases the unique counterfactual reasoning task—requiring the model to predict potential consequences if the ego vehicle were to take a different action (e.g., accelerating instead of decelerating). This design compels the model to not only identify the current state but also comprehend the underlying causal logic of the interaction.

To ensure data rigor while evaluating high-level cognitive capabilities, this study adopts an asymmetric design strategy for the partitioning of the IEDD-VQA dataset. Given that perception and recognition (L1), behavior description (L2), and physical quantization (L3) correspond directly to trajectory ground truths and possess unique, deterministic labels, they constitute the core of the training set. This approach aims to establish the model's precise cognition of the physical laws within driving scenarios through strong supervisory signals.

In contrast, counterfactual reasoning (L4) is inherently an open-ended causal prediction that lacks a unique physical ground truth as a supervisory signal. To prevent the introduction of uncertainty noise or the induction of model hallucinations during the training phase, the L4 reasoning task is reserved exclusively for the test set. This design establishes a challenging Out-of-Distribution (OOD) evaluation benchmark: it examines whether a model, after learning only objective physical facts ("what is involved" and "what happens"), can zero-shot derive hypothetical consequences based on its acquired physical common sense. Consequently, the test set contains complete L1–L4 task chains for 100 typical interaction scenarios, whereas the training set is strictly limited to objective factual descriptions (L1–L3).

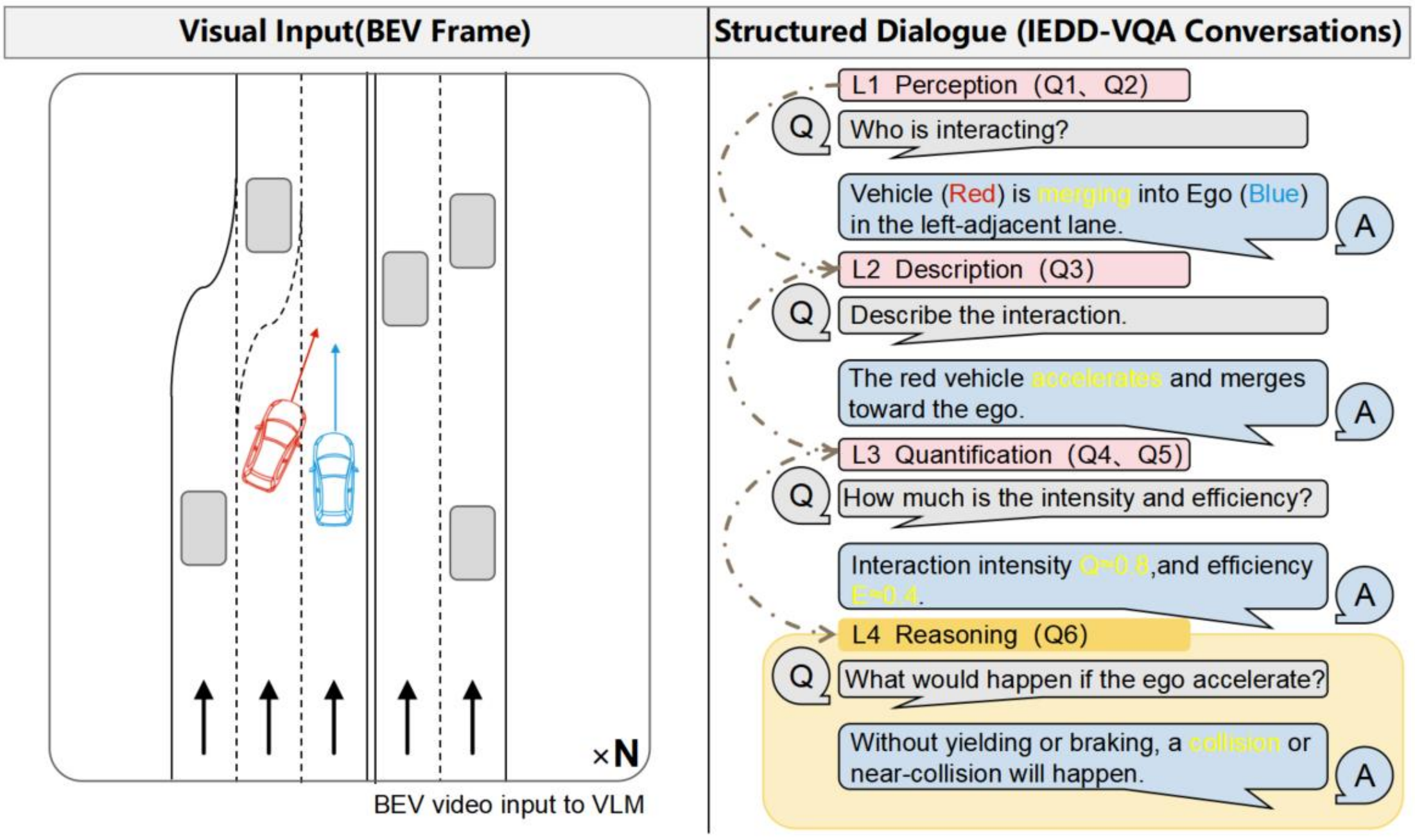

**Fig. 6** Task definitions and content examples of the IEDD-VQA dataset.

# Technical Validation

To comprehensively evaluate the fine-grained understanding and reasoning capabilities of VLMs in the specialized domain of autonomous driving interactions, this study establishes a hierarchical and multi-dimensional evaluation framework. Mainstream models are rigorously assessed under three distinct experimental settings. This section elaborates on the experimental setup, the evaluation metric system, and the design rationale for the comparative experiments.

## Experimental design

This section elaborates on the specific implementation details and evaluation criteria of the experiments. To ensure a fair comparison among VLM models with different architectures under a unified benchmark and to eliminate interference from data format discrepancies, the modal representation and preprocessing workflows for input data were first standardized. Subsequently, a hierarchical evaluation metric system was established across four dimensions—perception, description, quantization, and reasoning—aiming to comprehensively quantify the models' overall performance in interactive driving tasks.

(1) ***Input representation and data preprocessing***

This experiment constructs a high-quality evaluation set comprising 100 typical interaction scenarios based on the IEDD-VQA test set. Considering the varying compatibility of different VLMs with video inputs, a unified multi-frame image sampling strategy is adopted to ensure the fairness and robustness of the evaluation. For each interaction segment:

•Temporal Sampling: Six key frames are uniformly extracted from the video to cover the entire temporal progression of the traffic interaction.

• Image Preprocessing: All images are resized to a fixed resolution ($512 \times 512$) and Base64-encoded to preserve essential vehicle texture details while effectively managing token consumption.

• Context Construction: The 6-frame image sequence is utilized as a comprehensive visual context, provided as input only in the initial round of multi-turn dialogues. Subsequent question-answering relies entirely on the model's context retention capability.

(2) ***Hierarchical and multi-dimensional evaluation metric system***

The design of the evaluation criteria follows the principle that interactive driving decisions should be assessed not only by perceptual correctness, but also by their rationality in terms of safety, efficiency, interaction consistency, and interpretability. Recent studies on scenario-based decision-making evaluation have emphasized that rational autonomous driving behavior should be examined from multiple dimensions, including driving safety, driving efficiency, interaction-aware cooperation, and interpretability[59]. Inspired by these rational criteria, the proposed L1–L4 hierarchy is designed to progressively evaluate whether a VLM can perceive interaction objects, describe agent behaviors, quantify physical interaction states, and reason about potential consequences under counterfactual actions. In this sense, the Interaction Efficiency metric reflects the efficiency and smoothness of interaction execution, while the Counterfactual Reasoning score evaluates whether the model can provide causally consistent and safety-aware explanations for alternative decisions. This design situates IEDD-VQA within the broader evaluation landscape of interactive autonomous driving, while adapting the criteria to the specific requirements of vision-language reasoning.

To comprehensively evaluate the perception, description, quantitative analysis, and reasoning capabilities of VLMs in autonomous driving scenarios, this study constructs a hierarchical evaluation framework. For each test sample, the evaluation system comprises four progressive levels and incorporates the LLM-as-a-Judge paradigm. A high-performance large language model, GLM-4.7, is utilized as the evaluator to score the physical rationality and safety of the generated content. These scores are ultimately aggregated into a Weighted Integrated Score (WIS).

(2.1) ***Level 1: Perception and Identification*** This level assesses the model's fundamental perception of key elements in traffic scenarios using two metrics:

- Object ID Intersection over Union (Obj_IoU): For the identification of key vehicles, the set of IDs extracted from the model's predicted text is compared with the ground-truth set to calculate the Intersection over Union (IoU).
- Interaction Accuracy (Int_Acc): This evaluates the accuracy of the model in classifying vehicle interaction types.

(2.2) ***Level 2: Action Description Quality*** For the natural language description of driving behaviors, a dual evaluation standard of "semantic scoring + lexical overlap" is adopted:

- LLM Action Semantic Score (Act_Sem): The evaluator model assigns a score from 0 to 10 based on semantic consistency between the predicted text and the ground truth, focusing on "action correctness" and "hallucination-free generation." The score for this level is the normalized evaluator rating.
- Auxiliary Metric (ROUGE-L): The ROUGE-L (Recall-Oriented Understudy for Gisting Evaluation) score is calculated between the predicted and reference texts. Specifically, the F-measure based on the Longest Common Subsequence (LCS) is reported to quantify the lexical overlap between the generated text and the standard answer, compensating for the potential lack of granularity in pure semantic scoring.

(2.3) ***Level 3: Quantitative and Logical Analysis*** This level examines the model's numerical estimation capabilities and logical consistency regarding interaction intensity and efficiency parameters, primarily featuring:

- Mean Absolute Error (MAE): To intuitively measure the accuracy of physical parameter estimation, the average of the absolute differences between the predicted and ground-truth values across all test samples is calculated. This metric directly reflects the average deviation magnitude of the model at the physical scale; a lower value indicates more precise physical perception by the model.

$$MAE=\frac{1}{N}\sum_{i=1}^{N}\left|a_{gt}^{i}-a_{pred}^{i}\right| \tag{11}$$

- Logical Consistency (Log_Acc): This is a binary metric used to verify whether the model's judgment regarding the relative magnitudes of two physical quantities aligns with the ground truth. Rather than relying on specific numerical precision, this metric focuses on evaluating the model's logical reasoning capability regarding the interrelationships between objects in a scene.

When calculating the final score for this level (L3), to integrate the non-normalized MAE

with the probabilistic logical consistency, the MAE is first mapped to a normalized score in the range of [0, 1] based on a tolerance threshold of 0.5. This score is then averaged with the logical consistency accuracy to represent the overall performance of this level.

(2.4) ***Level 4: Counterfactual Reasoning*** This level requires the model to address complex reasoning queries, such as predicting the consequences of alternative actions.

• Reasoning Quality Score (Reas_Score): The evaluator model focuses on the causal logic and safety boundary awareness of the reasoning process, with scores 0 – 10, then normalized to the [0,1] for WIS.

• Auxiliary Metric (ROUGE-L): Similar to Level 2, the F1-measure of ROUGE-L is calculated to evaluate the similarity between the generated reasoning text and expert-annotated answers regarding key terminology and phrasing.

(2.5) ***Weighted Integrated Score (WIS)*** To derive a single-point metric reflecting the model's comprehensive capability, the performance across the four levels is integrated via a weighted sum. First, the metrics for each level are normalized into scores L1 through L4 within the [0, 1] range. Considering that safety-critical long-tail reasoning is paramount in autonomous driving tasks, a higher weight is assigned to the reasoning level in the final WIS calculation:

$$WIS=0.2 \cdot L1+0.2 \cdot L2+0.2 \cdot L3+0.4 \cdot L4 \tag{12}$$

The WIS ranges from 0 to 1, where a higher score indicates superior overall performance in autonomous driving interaction tasks.

(3) ***Human validation of the LLM-as-a-Judge evaluation***

To examine the reliability of the automated L4 counterfactual reasoning evaluation and to reduce the potential bias introduced by the LLM-as-a-Judge paradigm, we conducted a small-scale human validation study. Specifically, we randomly sampled 100 counterfactual QA responses from the L4 test results, while ensuring coverage of different interaction types, including car-following, merging, crossing, and head-on scenarios. Three researchers with backgrounds in autonomous driving and traffic safety were invited to independently score each response on a 0–10 scale. The annotators were blinded to the model identity and the GLM-4.7 scores.

The human evaluation followed the same criteria used by the automated judge, focusing on causal consistency, physical plausibility, safety awareness, scene-specific grounding, and hallucination avoidance. The final human score for each response was obtained by averaging the scores from the three annotators. We then calculated the Pearson correlation, Spearman rank correlation, and mean absolute error between the GLM-4.7 scores and the averaged human scores.

The results in Table 4 show that the GLM-4.7 scores are strongly aligned with human expert judgments, with a Pearson correlation of 0.8±0.03, a Spearman correlation of 0.78±0.04, and a mean absolute deviation of 1.22±0.12. These results suggest that although LLM-as-a-Judge evaluation may still contain model-specific preferences, GLM-4.7 provides a reliable approximation of expert judgment for large-scale counterfactual reasoning evaluation in IEDD-VQA.

| | Pearson correlation | Spearman correlation | MAE |
|---|---|---|---|
| Expert1 | 0.81 | 0.78 | 1.21 |
| Expert2 | 0.79 | 0.75 | 1.35 |
| Expert3 | 0.84 | 0.82 | 1.11 |

| Average | 0.81±0.03 | 0.78±0.04 | 1.22±0.12 |
|---|---|---|---|

**Table 4.** Human validation of the GLM-4.7-based LLM-as-a-Judge evaluation.

## Benchmark evaluation and analysis of mainstream VLMs

Based on the aforementioned hierarchical evaluation framework, this study selects several representative VLMs for a systematic evaluation. The experiments are designed to address two core questions: first, do general-purpose mainstream VLMs possess the fundamental capability to understand complex interaction scenarios without domain-specific fine-tuning in the autonomous driving field? Second, can prompt engineering strategies, such as CoT, effectively unlock the potential of these models in physical quantization and logical reasoning? This section begins by assessing zero-shot benchmark performance to analyze the distribution of strengths and weaknesses across different capability levels, followed by an exploration of the impact of prompting strategies on model performance.

(1) ***Zero-shot performance evaluation***

To establish the baseline performance of the IEDD-VQA dataset, ten representative VLMs were selected for zero-shot evaluation. The evaluation subjects encompass a diverse range of architectures, from lightweight to large-scale parameters.The evaluated models include recently released VLMs such as Kimi K2.5[60] and Gemini 2.5 Flash[61], together with other representative open-source and closed-source models. Notably, this experiment includes three open-source models (Llama-4-Maverick, Qwen2.5-VL-7B, and GLM-4.6V), while the remainder are closed-source commercial models. As illustrated in Table 5 and Fig. 7A, the experimental results show clear performance differences across models, with several open-source models achieving competitive scores.

| Model | Overall | Perception (L1) | Perception (L1) | Description (L2) | Description (L2) | Quant. (L3) | Quant. (L3) | Reasoning (L4) | Reasoning (L4) |
|---|---|---|---|---|---|---|---|---|---|
| | WIS | Obj IoU | Int Acc | Act Sem | ROUGE | MAE | Log Acc | Reas Score | ROUGE |
| seed-1.6-flash | 0.1348 | 0.205 | 0.25 | 0.35 | 0.3048 | 1358.1 | 0.21 | 1.53 | 0.1529 |
| kimi-k2.5[60] | 0.0964 | 0.2514 | 0.31 | 0.23 | 0.2495 | 650.8 | 0.21 | 0.36 | 0.1448 |
| llama-4-maverick | 0.342 | 0.2304 | 0.25 | 0.28 | 0.2311 | 714.3 | 0.43 | 4.93 | 0.0921 |
| nova-2-lite-v1 | 0.1876 | 0.334 | 0.25 | 0.23 | 0.2246 | 1084.7 | 0.24 | 2.52 | 0.1019 |
| glm-4.6v | 0.1696 | 0.3592 | 0.27 | 0.27 | 0.2866 | 1359.4 | 0.25 | 1.91 | 0.156 |
| claude-3-haiku | 0.0934 | 0.3183 | 0.3 | 0.73 | 0.2376 | 1283.3 | 0.11 | 0.14 | 0.1222 |
| gemini-2.5-flash[61] | 0.2091 | 0.3767 | 0.29 | 0.34 | 0.356 | 1352.7 | 0.39 | 2.42 | 0.1553 |
| gpt-4o | 0.2004 | 0.2958 | 0.35 | 0.26 | 0.2694 | 1358.1 | 0.21 | 2.74 | 0.1262 |
| grok-4.1-fa | 0.16 | 0.2974 | 0.26 | 0.14 | 0.2257 | 136 | 0.44 | 1.47 | 0.1282 |

| st | 12 | | | | | 0.6 | | | |
|---|---|---|---|---|---|---|---|---|---|
| qwen2.5-vl-7b | 0.2747 | 0.3795 | 0.24 | 0.31 | 0.2599 | 1855.5 | 0.15 | 4.66 | 0.1195 |

**Table 5.** Zero-shot evaluation results of ten mainstream VLMs on IEDD-VQA.

In terms of the Weighted Integrated Score (WIS), the open-source model Llama-4-Maverick ranked first with a score of 0.342, achieving the highest WIS among the evaluated models. It is followed by Qwen2.5-VL-7B (WIS=0.275), another open-source model, and the closed-source model Gemini-2.5-flash (WIS=0.209). In contrast, widely recognized closed-source flagship models such as GPT-4o (WIS=0.200) and Claude-3-Haiku (WIS=0.093) did not hold an advantage in zero-shot performance within this specialized domain. The radar charts in Fig. 7A intuitively illustrate this trend: the three open-source models in the top row generally yield larger envelope areas across the perception (L1), description (L2), and reasoning (L4) dimensions compared to or on par with the closed-source models in the second row. These results provide evidence that in the specific vertical domain of autonomous driving interactions, well-optimized open-source models have the potential to rival or even exceed the performance of top-tier closed-source models.

However, all models exhibited a significant bottleneck in the quantitative analysis (L3) dimension. As indicated in Table 5, in the absence of fine-tuning, the Mean Absolute Error (MAE) for all models is generally extremely high (e.g., 1855.5 for Qwen2.5-VL-7B and 1358.1 for GPT-4o). This suggests that although general-purpose VLMs possess strong semantic comprehension, extracting precise physical metrics directly from BEV videos remains a critical challenge. General models struggle to establish a direct and accurate mapping between visual features and physical numerical values.

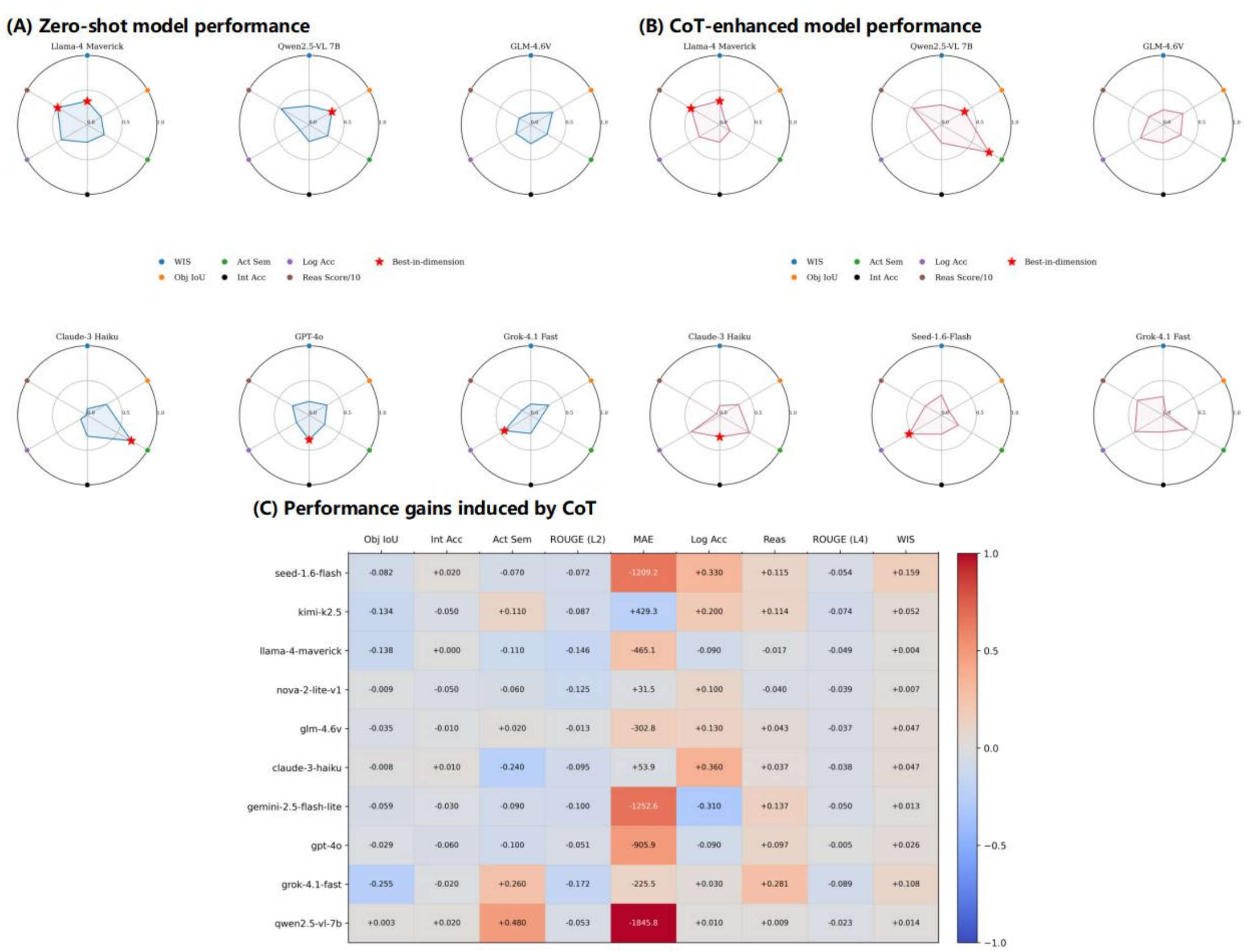

**Fig. 7** Evaluation of mainstream VLMs under zero-shot and CoT settings.

(2) ***Activation of logical reasoning capabilities via CoT***

To investigate the impact of in-context learning and logical guidance on model performance, we re-evaluated the models using a CoT strategy. The results, summarized in Table 6 and Fig. 7B, demonstrate that the CoT strategy significantly activates the latent logical reasoning capabilities of certain models, particularly open-source models with strong foundational performance.

The most notable improvement occurred in Qwen2.5-VL-7B. With the introduction of CoT, the model's WIS rose to 0.289. More impressively, its physical quantity estimation error (MAE) in the L3 level decreased substantially from a baseline of 1855.5 to 9.73. The heat map in Fig. 7C further corroborates this, with deep red regions highlighting the marked improvement in MAE. This phenomenon suggests that the interactive logical structures embedded in the IEDD dataset can be effectively extracted through proper prompt engineering. For models with robust instruction-following capabilities, CoT successfully shifts their focus from pure visual perception to step-by-step reasoning based on physical common sense, achieving a substantial improvement in numerical estimation.

However, CoT did not yield positive gains across all models. Certain models, such as Llama-4-Maverick and Claude-3-Haiku, experienced negative growth in L2 description scores (e.g., Llama-4's Act_Sem decreased by 0.11). This may be because the long-text reasoning process introduced by CoT partially interferes with the generation of concise driving action descriptions, leading to "semantic drift."

| Model | Overall | Perception (L1) | Perception (L1) | Description (L2) | Description (L2) | Quant. (L3) | Quant. (L3) | Reasoning (L4) | Reasoning (L4) |
|---|---|---|---|---|---|---|---|---|---|
| | WIS | Obj IoU | Int Acc | Act Sem | ROUGE | MAE | Log Acc | Reas Score | ROUGE |
| seed-1.6-flash | 0.2941 | 0.123 | 0.27 | 0.28 | 0.2326 | 148.932 | 0.54 | 2.68 | 0.0992 |
| kimi-k2.5[60] | 0.1483 | 0.1177 | 0.26 | 0.34 | 0.1629 | 1080.1333 | 0.41 | 1.5 | 0.0711 |
| llama-4-maverick | 0.346 | 0.0923 | 0.25 | 0.17 | 0.085 | 249.24 | 0.34 | 4.76 | 0.0435 |
| nova-2-lite-v1 | 0.195 | 0.3247 | 0.2 | 0.17 | 0.0999 | 1116.2373 | 0.34 | 2.12 | 0.0632 |
| glm-4.6v | 0.2171 | 0.324 | 0.26 | 0.29 | 0.2737 | 1056.5553 | 0.38 | 2.34 | 0.119 |
| claude-3-haiku | 0.1407 | 0.3108 | 0.31 | 0.49 | 0.1431 | 1337.182 | 0.47 | 0.51 | 0.084 |
| gemini-2.5-flash[61] | 0.2226 | 0.3181 | 0.26 | 0.25 | 0.2561 | 100.1293 | 0.08 | 3.79 | 0.105 |
| gpt-4o | 0.2262 | 0.2671 | 0.29 | 0.16 | 0.2182 | 452.194 | 0.12 | 3.71 | 0.1211 |
| grok-4.1-fast | 0.2692 | 0.0429 | 0.24 | 0.4 | 0.0537 | 1135.1347 | 0.47 | 4.28 | 0.0391 |
| qwen2.5-vl-7b | 0.2887 | 0.3823 | 0.26 | 0.79 | 0.2071 | 9.7333 | 0.16 | 4.75 | 0.096 |

**Table 6.** Model performance evaluation results after introducing the CoT strategy.

## Domain-adaptive fine-tuning and ablation studies

To validate the value of the IEDD dataset in domain adaptation and explore methods for transforming general-purpose VLMs into expert-level driving interaction models, we conducted Low-Rank Adaptation (LoRA)[62] fine-tuning experiments on the top-performing open-source Qwen2.5-VL-7B based on the IEDD-VQA training set.

The experimental implementation was conducted using the PyTorch 2.6.0 and Parameter-Efficient Fine-Tuning (PEFT) 0.12.0 frameworks. We utilized two Xiyun C500 GPUs and employed the LoRA efficient fine-tuning strategy, with Qwen2.5-VL-7B-Instruct serving as the base model. To optimize training stability under constrained computational resources, the total training batch size was set to 128, achieved through 64 gradient accumulation steps. The AdamW optimizer was adopted with an initial learning rate of 5e-05, supplemented by a cosine decay scheduling strategy and a warmup ratio of 0.05 to prevent gradient oscillations in the early stages of training. The model underwent training for 2 epochs on the IEDD-VQA training set, with the final validation loss converging to 0.0993. The loss convergence curve is depicted in Fig. 8A.

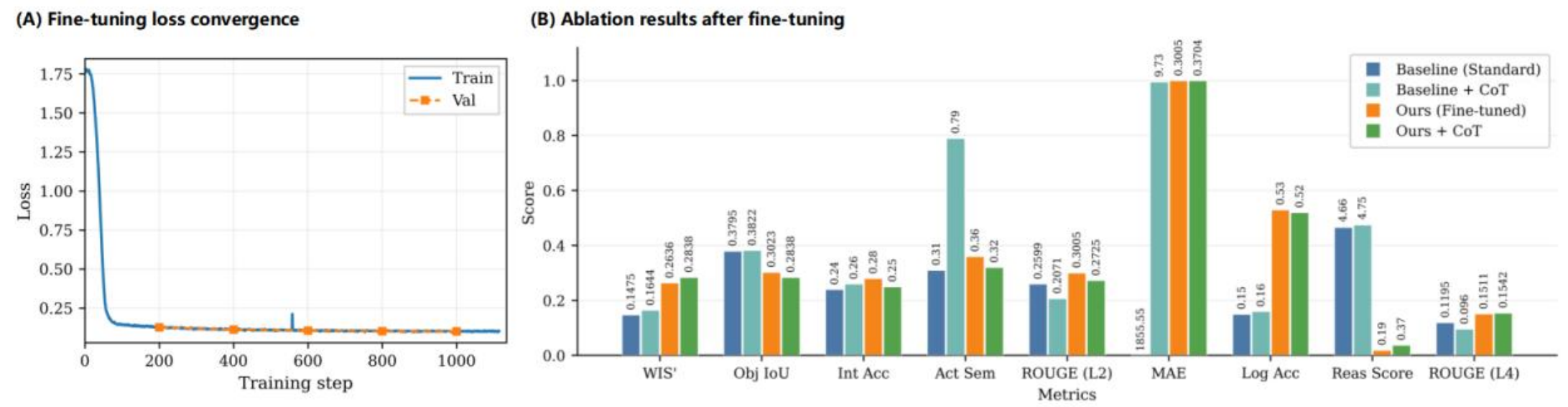

**Fig. 8** Fine-tuning convergence and ablation analysis.

Since the fine-tuning dataset solely covers perception (L1), description (L2), and quantification (L3) tasks and lacks targeted training for counterfactual reasoning (L4), this section employs the revised comprehensive metric WIS' for evaluation. Meanwhile, the L4 metric is retained to monitor the model's performance on OOD tasks.

$$WIS' = \frac{(L1+L2+L3)}{3} \tag{13}$$

| Model | Overall | Perception (L1) | Perception (L1) | Description (L2) | Description (L2) | Quant. (L3) | Quant. (L3) | Reasoning (L4) | Reasoning (L4) |
|---|---|---|---|---|---|---|---|---|---|
| | WIS' | Obj IoU | Int Acc | Act Sem | ROUGE | MAE | Log Acc | Reas Score | ROUGE |
| Baseline (Standard) | 0.1475 | 0.3795 | 0.24 | 0.31 | 0.2599 | 1855.55 | 0.15 | 4.66 | 0.1195 |
| Baseline + CoT | 0.1644 | 0.3822 | 0.26 | 0.79 | 0.2071 | 9.73 | 0.16 | 4.75 | 0.096 |
| Ours (Fine-tuned) | 0.2636 | 0.3023 | 0.28 | 0.36 | 0.3005 | 0.3036 | 0.53 | 0.19 | 0.1511 |
| Ours + CoT | 0.2336 | 0.2838 | 0.25 | 0.32 | 0.2725 | 0.3704 | 0.52 | 0.37 | 0.1542 |

**Table 7.** Performance ablation study on in-distribution and out-of-distribution tasks before

and after fine-tuning.

The experimental results are presented in Table 7 and Fig. 8B. After fine-tuning on the IEDD-VQA dataset, the model (Ours) achieved significant performance gains across multiple core metrics, validating the effectiveness of domain-specific adaptation training:

• Significant Enhancement in Overall Performance: The revised weighted integrated score (WIS') of the fine-tuned model increased from 0.1475 (Baseline) to 0.2636, representing a 78.7% improvement. This result indicates that through targeted instruction fine-tuning, the model successfully bridged the domain gap between general-purpose scenarios and autonomous driving interactions, establishing stronger adaptability for in-distribution tasks.

• Precision in Physical Perception and Quantification: The most decisive breakthrough occurred at the L3 quantification level. Following fine-tuning, the Mean Absolute Error (MAE) converged sharply from an initial 1855.55 to 0.3036, while the logical accuracy (Log Acc) significantly improved from 0.15 to 0.53. This implies that the model not only mastered the semantic description of interactive scenarios but also accurately learned an internal representation mechanism that maps visual features to physical parameters such as speed and intensity, overcoming the inherent limitations of general-purpose models in numerical estimation.

• Alignment of Action Semantics and Reversal of CoT Effects: At the L2 description level, the action semantics score (Act Sem) steadily rose from 0.31 to 0.36, indicating that the generated descriptions of driving behavior aligned more closely with professional terminology. Notably, comparing the results of Ours (Fine-tuned) with Ours + CoT reveals that after sufficient fine-tuning, introducing CoT actually caused the WIS' to drop from 0.2636 back to 0.2336. This suggests that fine-tuning on the IEDD dataset has internalized complex interaction reasoning logic into the model's task-specific response patterns; consequently, additional explicit reasoning steps may constitute information redundancy or interference, thereby reducing the quality of end-to-end outputs.

• Trade-off Between Domain Specialization and General Capability: However, the degradation in OOD reasoning effect brought by fine-tuning cannot be ignored. On the L4 counterfactual reasoning task, which was not included in the fine-tuning data, the reasoning score decreased substantially from a baseline of 4.66 to 0.19. This phenomenon underscores the limitations of LoRA fine-tuning in strengthening domain-specific capabilities: while the model becomes highly adapted to the specific instruction formats and physical ground truths of IEDD-VQA, it sacrifices the general logical reasoning capabilities and generalization advantages for open-ended unknown scenarios acquired during the pre-training phase.

In summary, the IEDD dataset effectively transforms a general VLM into a "domain-adapted model" focused on interactive perception and quantification, albeit at the cost of certain general cognitive abilities. This suggests that future research should incorporate broader task replay or regularization strategies during instruction fine-tuning to maintain robustness on OOD data while pursuing peak specialized performance.

**Limitations.** Although IEDD provides a large-scale benchmark for interaction-oriented VLA learning, several limitations remain. First, the BEV-only modality mainly captures trajectory-level and topology-level interaction structures, but cannot fully represent appearance-level micro-semantics, such as pedestrian gaze, gestures, brake lights, and turn signals, which are important for human-like intent prediction. Second, IEDD is constructed from heterogeneous public trajectory datasets. Despite trajectory standardization, differences in sensing configurations, map quality, traffic rules, sampling frequency, and annotation formats may still introduce domain bias. Third, the interaction mining, quantification, and VQA construction processes rely on predefined thresholds, heuristic rules, category-adaptive metric weights, and rule-constrained language templates. These designs improve scalability and factual consistency, but may overlook ambiguous strategic negotiation, implicit intentions, and diverse human-like explanations. Future work will explore world-model-based generation to enrich BEV trajectory reconstruction with physically plausible raw-sensory details, such as local visual contexts, multi-view observations, and fine-grained semantic cues, while preserving the global interaction consistency derived from real trajectories.

# Data availability

The derived IEDD and IEDD-VQA records generated in this study are deposited in Zenodo (https://doi.org/10.5281/zenodo.18742437)[58]. The repository contains derived interaction annotations, structured semantic records, VQA metadata, and benchmark evaluation files. The raw source datasets are not redistributed. Users should obtain the original Waymo Open Motion Dataset, nuPlan, Lyft Level 5, INTERACTION, and SIND records from the respective data providers under their original licenses or terms of use. Source-dataset provenance fields are retained in the released annotations to allow users with authorized access to the source datasets to locate the corresponding source records and reproduce the processing pipeline.

# Code availability

The codes for generating IEDD-VQA and conducting benchmark evaluation are documented and publicly available at https://github.com/egik-von/IEDD. We provide four tools to help users construct and evaluate IEDD-VQA:

• IEDD-traj2VisAct.py renders BEV vision clips from trajectory data and extracts corresponding action semantics from interaction segments.

• IEDD-2vqa.py converts interaction annotations, action semantics, and video paths into ShareGPT-format question-answering data for Q1–Q5.

• IEDD-traj2Q6.py appends counterfactual reasoning questions to the generated Q1–Q5 data and constructs the IEDD-VQA_test benchmark.

• IEDD-benchmark.py runs benchmark evaluation on IEDD-VQA_test and records the model responses and evaluation results.

The data generation pipeline begins with IEDD-traj2VisAct.py, which uses the prepared trajectory cache and IEDD CSV annotations to generate BEV videos and action semantic files. These outputs are then used by IEDD-2vqa.py to construct ShareGPT-format VQA samples covering perception, behavior description, and physical quantification tasks. Next, IEDD-traj2Q6.py adds counterfactual reasoning questions to form the final IEDD-VQA_test benchmark. Finally, IEDD-benchmark.py evaluates selected VLMs on the generated benchmark and saves the corresponding results for further analysis.

# Funding

This work is jointly sponsored by the National Natural Science Foundation of China (Grant No. 52325212) and the National Key Research and Development Program of China (Grant No. 2024YFB2505704).